\documentclass{article}

\usepackage{csquotes}

\usepackage{microtype}
\usepackage{graphicx}
\usepackage{enumitem}

\usepackage{booktabs} % for professional tables

\usepackage[belowskip=-10pt,aboveskip=0pt]{subcaption}
\usepackage{amsmath,amsfonts,bm}

\def\eqref#1{equation~\ref{#1}}
\def\1{\bm{1}}

\DeclareMathAlphabet{\mathsfit}{\encodingdefault}{\sfdefault}{m}{sl}
\SetMathAlphabet{\mathsfit}{bold}{\encodingdefault}{\sfdefault}{bx}{n}

\newcommand{\E}{\mathbb{E}}

\usepackage{hyperref}

\usepackage{url}
\usepackage{float, graphicx,times}

\usepackage{wrapfig}
\usepackage{xcolor}
\definecolor{brilliantlavender}{rgb}{0.96, 0.73, 0.90}
\definecolor{blizzardblue}{rgb}{0.67, 0.9, 0.93}
\definecolor{green}{rgb}{0.0, 0.49411764705882355, 0.058823529411764705}
\usepackage[most]{tcolorbox}
\tcbset{on line, 
        boxsep=2pt, left=0pt,right=0pt,top=0pt,bottom=0pt,
        colframe=white,colback=blizzardblue,  
        highlight math style={enhanced}
        }

\definecolor{up}{rgb}{0,0.6,.34}
\definecolor{up1}{rgb}{0.35,0.72,.25}
\definecolor{down}{rgb}{0.9,0.2,.2}
\definecolor{blue}{rgb}{0.358,0.564,0.675}
\definecolor{br}{rgb}{0.48,0.15,.03}
\definecolor{sb}{rgb}{0.529,0.807,0.9215}
\usepackage{bibunits}
\usepackage{multirow}
\usepackage{float}
\usepackage{graphicx,subcaption}
\usepackage{booktabs}       % professional-quality tables

\usepackage{hyperref}

\usepackage[accepted]{icml2021}

\icmltitlerunning{Mean Embeddings with Test-Time Data Augmentation for Ensembling of Representations}

\begin{document}

\twocolumn[
\icmltitle{
    Mean Embeddings with Test-Time Data Augmentation\\
    for Ensembling of Representations}



\icmlsetsymbol{equal}{*}

\begin{icmlauthorlist}
\icmlauthor{Arsenii Ashukha}{saic}
\icmlauthor{Andrei Atanov}{epfl}
\icmlauthor{Dmitry Vetrov}{saic}
\end{icmlauthorlist}

\icmlaffiliation{saic}{Samsung AI Center Moscow}
\icmlaffiliation{epfl}{Swiss Federal Institute of Technology Lausanne (EPFL), work done while at National Research University Higher School of Economics}
\icmlcorrespondingauthor{Arsenii Ashukha}{ars.ashuha@gmil.com}
\icmlcorrespondingauthor{Andrei Atanov}{ai.atanow@gmail.com}

\icmlkeywords{Machine Learning, ICML}

\vskip 0.3in
]



\printAffiliationsAndNotice{\icmlEqualContribution} % otherwise use the standard text.

\begin{abstract}
Averaging \emph{predictions} over a set of models---an ensemble---is widely used to improve predictive performance and uncertainty estimation of deep learning models. 
At the same time, many machine learning systems, such as search, matching, and recommendation systems, heavily rely on embeddings. 
Unfortunately, due to misalignment of features of independently trained models, embeddings, cannot be improved with a naive deep ensemble like approach.
In this work, we look at the ensembling of representations and propose mean embeddings with test-time augmentation (MeTTA) simple yet well-performing recipe for ensembling representations. 
Empirically we demonstrate that MeTTA significantly boosts the quality of linear evaluation on ImageNet for both supervised and self-supervised models.
Even more exciting, we draw connections between MeTTA, image retrieval, and transformation invariant models.
\newline
\newline
We believe that spreading the success of ensembles to inference higher-quality representations is the important step that will open many new applications of ensembling.
\end{abstract}

\section{Mean Embeddings}
Our goal is to utilize ensembles to improve quality of embedding produced by a neural network.
Conventional ensembling techniques e.g., deep ensembles \citep{lakshminarayanan2017simple} involve averaging predictions of several independently trained networks. Representations inside these networks are not coherent: activations that appear on the same positions in different networks respond to different input signals, thus cannot be averaged. Below we explain how to get around this obstacle.

% We consider a deep learning model $p(y\,|\,x, \mathrm{w})$ that is trained on dataset $(x_i, y_i) \in \mathcal{D}$ of size $K$. 
We consider a model $p(y\,|\,x, \mathrm{w})$ represented by neural network with parameters $\mathrm{w}$ that is trained on dataset $(x_i, y_i) \in \mathcal{D}$ of size $K$, where $x$ denotes an object, $y$ denotes a variable to be predicted, and $\mathrm{w}$ denotes parameters of the model. 
The model is a composition of transformations $f(x, \mathrm{w})=f^N_{\mathrm{w}^N} \circ \dots \circ f^1_{\mathrm{w}^1} (x)$. 
We denote intermediate representations or activations of $L$-th layer as $a^L(x; \mathrm{w})=f^L_{\mathrm{w}^L} \circ \dots \circ f^1_{\mathrm{w}^1} (x)$.
% We assume that the model is trained by optimizing a loss function $\mathcal{L}$ and used data augmentation during training $\hat{x}\sim p_{aug}(\hat{x}\,|\,x)$ and mini-batch based stochastic optimization 
We assume the model is trained by optimizing a loss function $\mathcal{L}$ with a stochastic optimization and using data augmentation $\hat{x}\sim p_{aug}(\hat{x}\,|\,x)$.
\begin{equation}
    \E_{\textcolor{white}{\hat{\mathcal{B}}}\!\!\!\!x \sim \mathcal{D}} \E_{\hat{x}\sim p_{aug}(\hat{x}\,|\,x)} \mathcal{L}(\hat{x}, \mathrm{w}) \to \min_{\mathrm{w}}.
\end{equation}

\begin{figure}[t!]
\centering
    \includegraphics[width=1\linewidth]{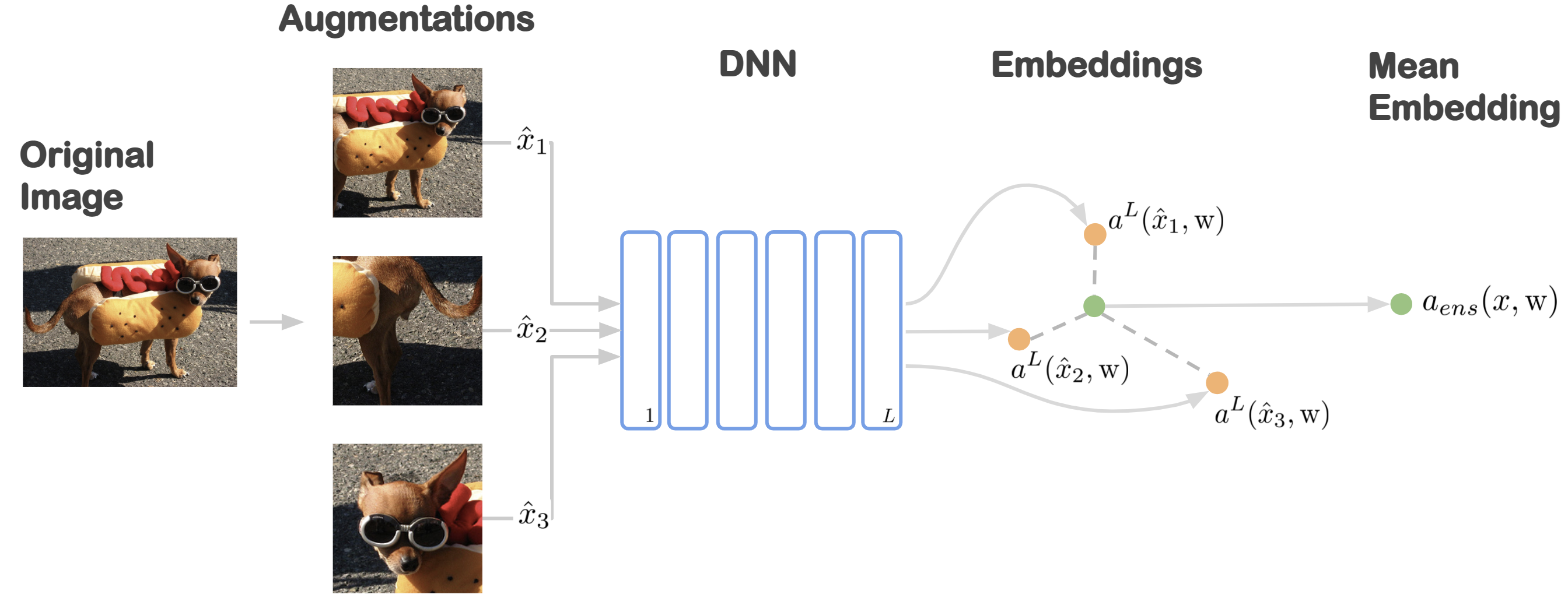}
    
\caption{
To produce a \emph{mean embedding} MeTTA averages activations of a network across different augmentations of an object. 
The MeTTA does not affect training phase and can be applied to a pre-trained network.
}
\label{fig:me}
\end{figure}
We propose \textbf{\emph{mean embeddings with TTA (MeTTA)}}---a simple method for representations ensembling.
The method averages representations $a^L(\,\cdot\,; \mathrm{w})$ of a single model over different augmentations of an object $x$
\begin{gather}
    a_{ens}(x; \mathrm{w}) =  \E_{\hat{x} \sim p_{aug}(\cdot \,|\,x)} a^L(\hat{x}; 
    \mathrm{w}) \cong \\
    \cong \frac{1}{S} \sum_{s=1}^{S} a^L(\hat{x}_s; 
    \mathrm{w}), \text{~where~~} \hat{x}_s \sim p_{aug}(\cdot \,|\,x),
\end{gather}
where $S$ is number of samples of augmentations for a single image, and $a_{ens}(x; \mathrm{w})$ is a mean embedding.
\begin{table*}[]
\centering
\resizebox{0.7\textwidth}{!}{% <------ Don't forget this %
\begin{tabular}{clcccccc}
\toprule
\multicolumn{5}{c}{} & Central crop & \multicolumn{2}{c}{Mean embeddings with TTA}\\ 
Problem & Model    & Width & SK & $\#$ Params (M) &  Embeddings & $N=10$ & $N=32$ \\
\midrule
Self-supervised & ResNet50 &   $1\times$  &  False &  $24$   & $71.7$ & $73.3$ {\scriptsize (\textcolor{up}{+$1.6\%$})} & $73.8$ {\scriptsize (\textcolor{up}{+$2.1\%$})}    \\
features&                   &   $1\times$  &  True  &  $35$   & $74.6$ & $75.8$ {\scriptsize (\textcolor{up}{+$1.2\%$})} & $76.2$ {\scriptsize (\textcolor{up}{+$1.6\%$})}   \\
\textcolor{gray}{(SimCLRv2)}&  ResNet101                &   $2\times$  &  False &  $170$  & $77.0$ & $78.1$ {\scriptsize (\textcolor{up}{+$1.1\%$})} & $78.5$ {\scriptsize (\textcolor{up}{+$1.5\%$})}  \\
&                           &   $2\times$  &  True  &  $257$  & $79.0$ & $79.8$ {\scriptsize (\textcolor{up}{+$0.8\%$})} & $79.9$ {\scriptsize (\textcolor{up}{+$0.9\%$})}\\
&  ResNet152                &   $3\times$  &  True  &  $795$  & $79.8$ & $80.3$ {\scriptsize (\textcolor{up}{+$0.4\%$})} & $80.7$ {\scriptsize (\textcolor{up}{+$0.9\%$})}\\
\cmidrule(lr){1-8}
Supervised & ResNet50       &   $1\times$  &  False &  $24$   & $76.6$ & $78.0$ {\scriptsize (\textcolor{up}{+$1.4\%$})} & $78.5$  {\scriptsize (\textcolor{up}{+$1.9\%$})}\\
features &                  &   $1\times$  &  True  &  $35$   & $78.5$ & $79.7$ {\scriptsize (\textcolor{up}{+$1.2\%$})} & $80.2$  {\scriptsize (\textcolor{up}{+$1.7\%$})}\\
&  ResNet101                &   $2\times$  &  False &  $170$  & $78.9$ & $80.2$ {\scriptsize (\textcolor{up}{+$1.3\%$})} & $80.6$  {\scriptsize (\textcolor{up}{+$1.7\%$})}\\
&                           &   $2\times$  &  True  &  $257$  & $80.1$ & $81.0$ {\scriptsize (\textcolor{up}{+$0.9\%$})} & $81.3$  {\scriptsize (\textcolor{up}{+$1.3\%$})}\\
&  ResNet152                &   $3\times$  &  True  &  $795$  & $80.5$ & $81.4$ {\scriptsize (\textcolor{up}{+$0.9\%$})} & $81.9$ {\scriptsize (\textcolor{up}{+$1.4\%$})}\\
\bottomrule
\end{tabular}}
\caption{Comparing of different methods of inference embeddings. 
Top-1 accuracy on ImageNet \citep{imagenet_cvpr09} for linear evaluation of embeddings with 100\% labels. 
For self-supervised features we used models that were pre-traoned with SimCLRv2 \cite{chen2020big}. 
We used both supervised and self-supervised pretrained models from {\small\url{https://github.com/google-research/simclr}} repository.
SK states for selective kernels \cite{li2019selective}.
For log-likelihood score see Fig. \ref{fig:qwe111}}
\label{res_clf}
\end{table*}
\definecolor{ultramarine}{rgb}{0.0, 0.0, 0.8}
% In other words, the MeTTA simply averages activations of the network over multiple augmentations of an object.
In contrast to a deep ensembles like approach\footnote{We discuss possible steps to DE-like approaches in Sec. 5}, global image representations, so called non-spatial activations, of a single network have no misalignment between different augmentations of a single image, so activations can be safely averaged.
The illustration of MeTTA is in Figure \ref{fig:me}.

% \begin{displayquote}
% \sl \textbf{\textcolor{gray}{``}}Ok, fine. But why would it work any better then a conventional inference?\textbf{\textcolor{gray}{''}}\\
% \rightline{--- Prof.~~~~~~~~~~~~~~~~~}
% \end{displayquote}
% \vspace{-0.5cm}
There are several reasons why MeTTA might work better then a conventional inference:
\begin{enumerate}
    \item \textbf{Transformation invariant representations~} When a model is trained with data augmentations, it is forced to be invariant to a label preserving transformations. In practice, though, the models only only partially are invariant to these transformations. Representations of different augmentations of one object may vary a lot, which over-complicates the embedding space and may significantly affect the model predictive performance. Mean embeddings, on the other hand, are invariant to such transformations by design.
    \item \textbf{Higher degree of parameter sharing during inference} Mean embeddings enjoys a higher degree of parameter sharing. It allows $N$ times more compute per parameter during inference, and potentially, use parameters more efficiently and boost the performance of the model. Mean embeddings, however, sacrifice computational complexity of inference that is a common drawback for ensembles.
    \item \textbf{Train-test distribution shift~} 
    A model usually sees augmented data during training and clean non-augmented data during test, which can potentially introduce a domain shift, and result in a loss of performance.
    MeTTA does not suffer from the shift because it has exactly the same distribution of data for training and inference. 
\end{enumerate}
% So that is my intuition. Does it sound convincing?

% \begin{displayquote}
% \sl \textbf{\textcolor{gray}{``}} Sort of. Fine, let's test the model at simple problem first. \textbf{\textcolor{gray}{''}}\\
% \rightline{--- Prof.~~~~~~~~~~~~~~~~~}
% \end{displayquote}

\section{Classification with Mean Embeddings}

We will use linear evaluation---a conventional protocol for evaluation of embedding that is widely used in  self-supervised learning~\cite{chen2020big}. The main idea is to train a linear classifier on top of the embeddings produced by a pretrained network, while weights of the network remain fixed. Then one can compare pretrained models based on test performance of the trained linear classifier.

\textbf{But how to train a model with mean embeddings?} More formally, we are supposed to train the following model $p(y\,|\,x)$ on top of the mean embeddings $a_{ens}(x, \mathrm{w})$:
\begin{multline}
  p(y\,|\,x)\,= \textsl{softmax}(\theta^T a_{\text{ens}}(x, \mathrm{w})) = \\
  = \textsl{softmax}(\theta^T \E_{\hat{x} \sim p_{\text{aug}}(\cdot \,|\,x)} a^L(\hat{x}; 
\mathrm{w})),
\label{model}
\end{multline}
by minimizing a cross entropy loss $$-\E_{(x, y) \sim \mathcal{D}}\log p(y\,|\,x) \to \min_{\theta}.$$ 
However, due to the intractable expectation inside $p(y\,|\,x)$ that is presented under the nonlinear function,
a sample-based gradients of this objective will be biased.
We, therefore, substitute the loss objective with the following upper bound, whose unbiased gradients can be estimated:
\begin{multline}
    -\log p(y\,|\,x)\!=\!-\log\textsl{sm}(\theta^T\!\tcbhighmath{\E_{\hat{x} \sim p_{\text{aug}}(\cdot \,|\,x)}} a^L(\hat{x}; \mathrm{w}))_y \leq\footnotemark\\
    \leq -\tcbhighmath{\E_{\hat{x} \sim p_{\text{aug}}(\cdot \,|\,x)}} ~\log\textsl{sm}(\theta^T a^L(\hat{x}; \mathrm{w}))_y \to \min_{\mathrm{w}, \theta},\label{upb}
\end{multline}
where {\sl sm} denotes {\sl softmax}. The equation \ref{upb} appears to be the conventional loss that is used for training a model with data augmentations. 
Thus, we can use a conventionally training with no additional changes needed.
\footnotetext{Jensen’s inequality: $f(\E x) \leq \E f(x)$ for concave $f=-\log \textit{softmax}$ \cite{boyd2004convex}.}
\stepcounter{footnote}

\textbf{MeTTA seems to help in practice!} Using mean embeddings show stable improvement over conventional central crop embedding for both supervised and self-supervised pre-trained models~(Table \ref{res_clf}). 

\definecolor{ultramarine}{rgb}{0.0, 0.0, 0.8}
\section{Understanding}

\textbf{Intuition~}
Deep learning models are usually trained to raise predictions that are invariant to data augmentations.
This principle is heavily used by supervised learning, and especially self-supervised learning methods based on contrastive losses e.g., SimCLR. 
% During training, a network aims to pull together embeddings of individual augmentations.

\begin{figure}[t!]
\centering
    \begin{subfigure}{1\linewidth}
    \centering
        \includegraphics[width=0.95\linewidth]{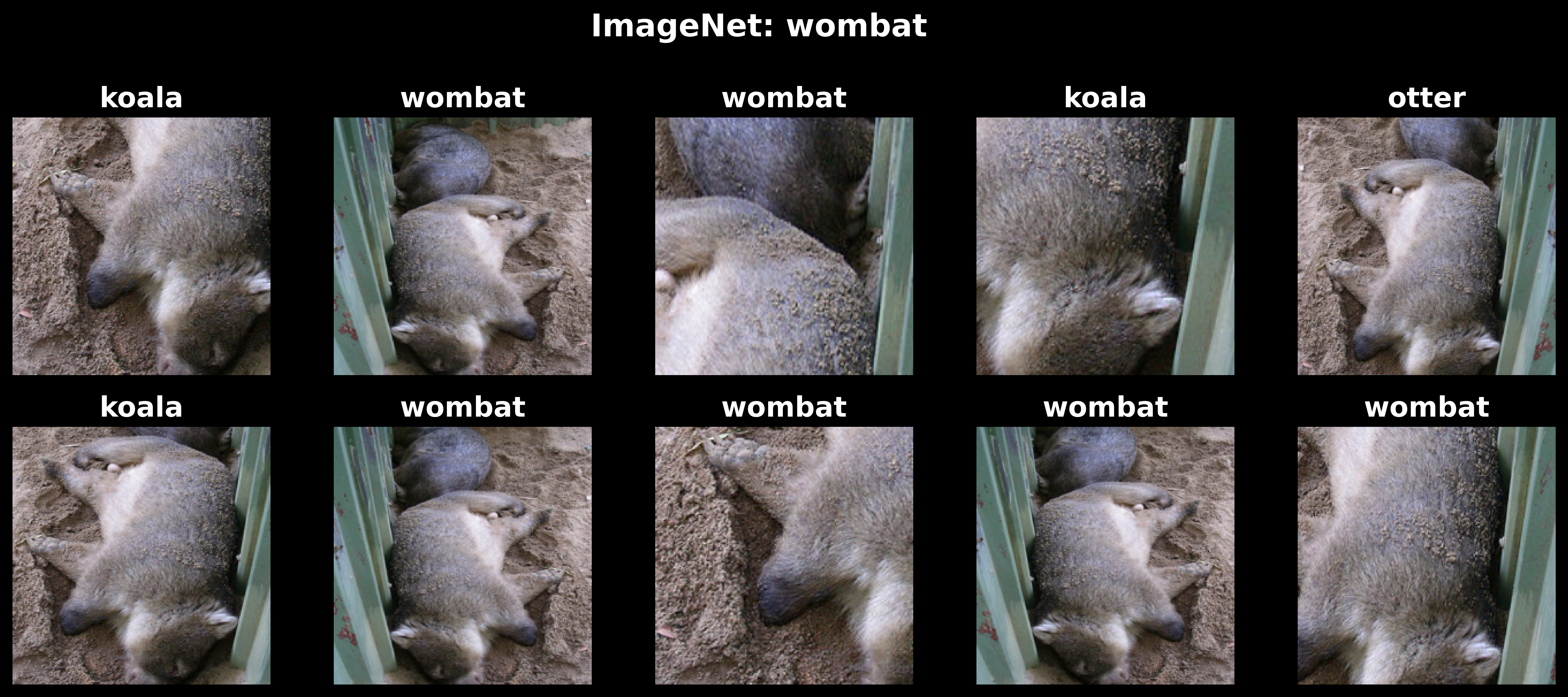}
        \label{ll3}
    \end{subfigure}
    
    \begin{subfigure}{1\linewidth}
    \centering
        \includegraphics[width=0.95\linewidth]{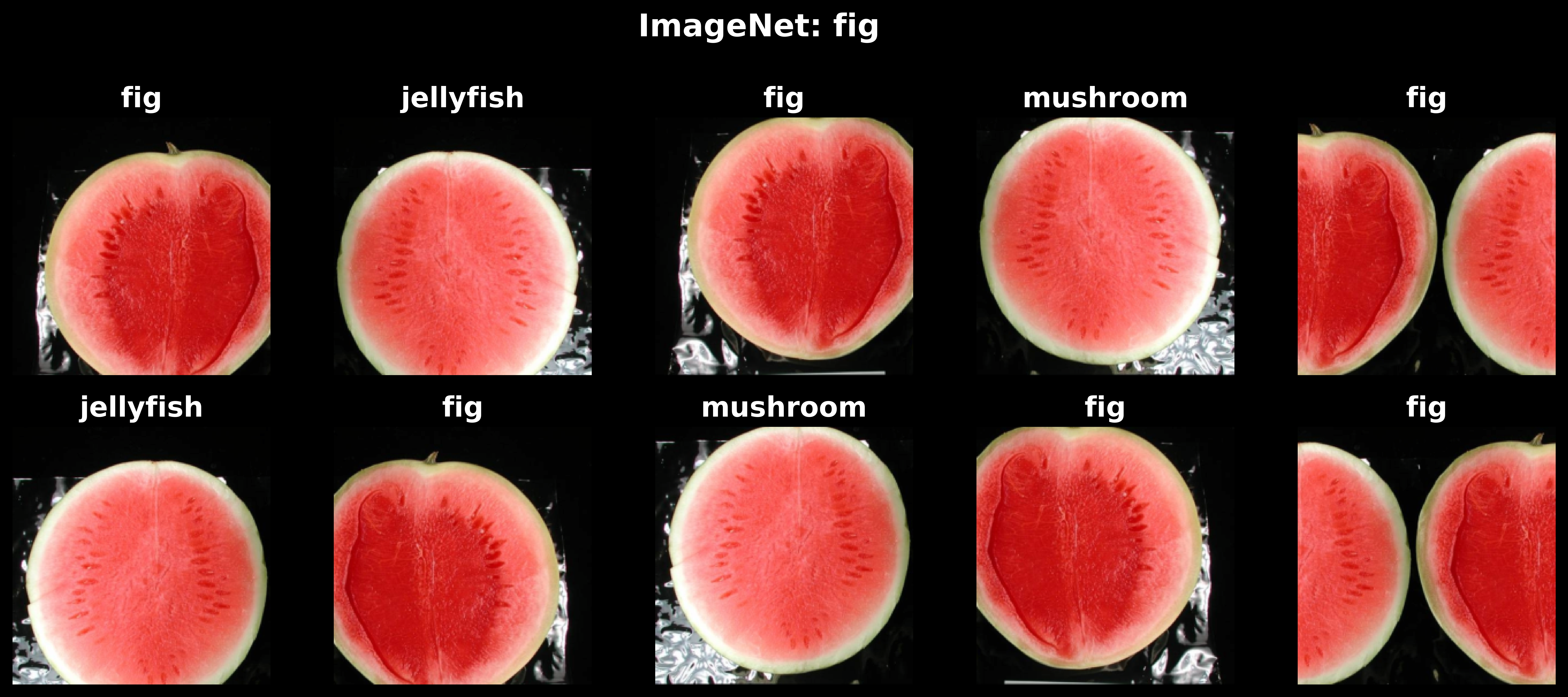}
        \label{ll3}
    \end{subfigure}
        
    \begin{subfigure}{1\linewidth}
    \centering
        \includegraphics[width=0.95\linewidth]{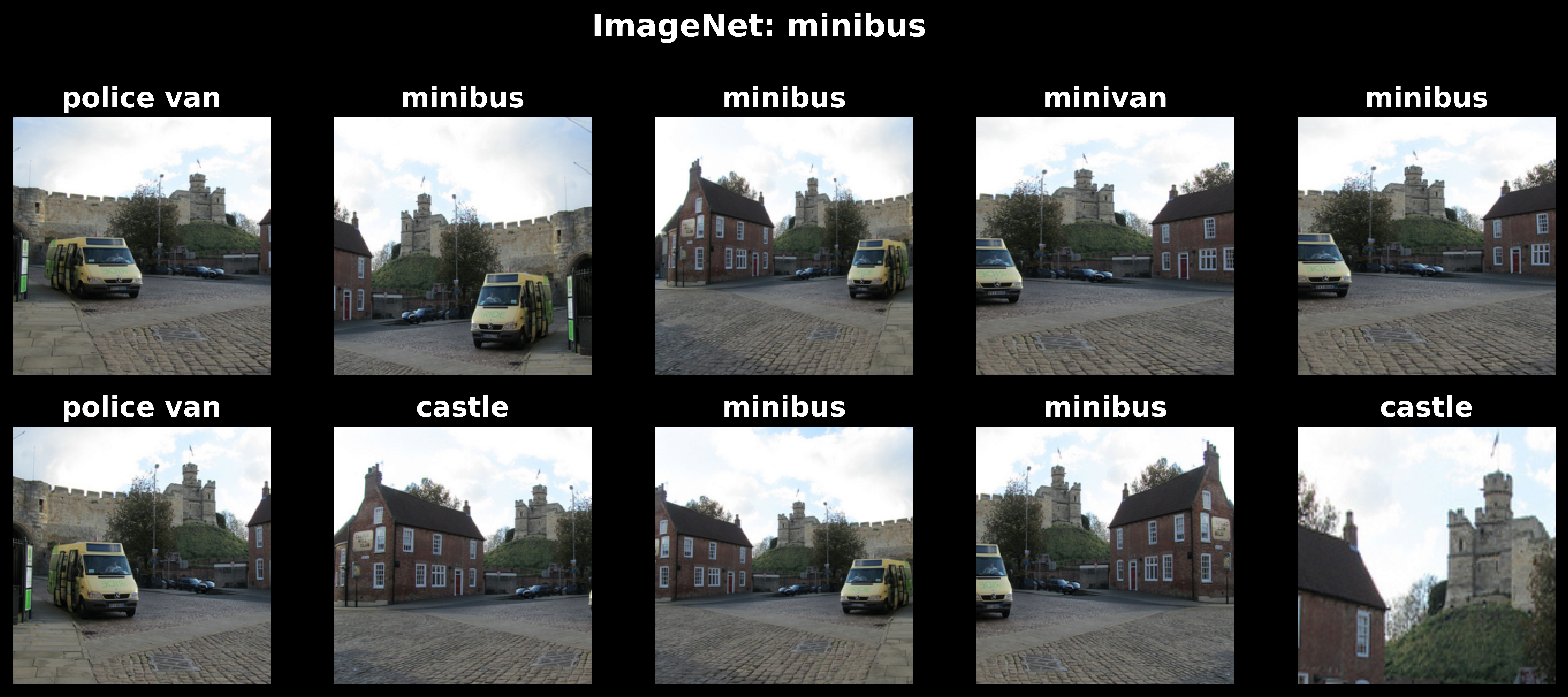}
        \label{ll3}
    \end{subfigure}
    
        \begin{subfigure}{1\linewidth}
    \centering
        \includegraphics[width=0.95\linewidth]{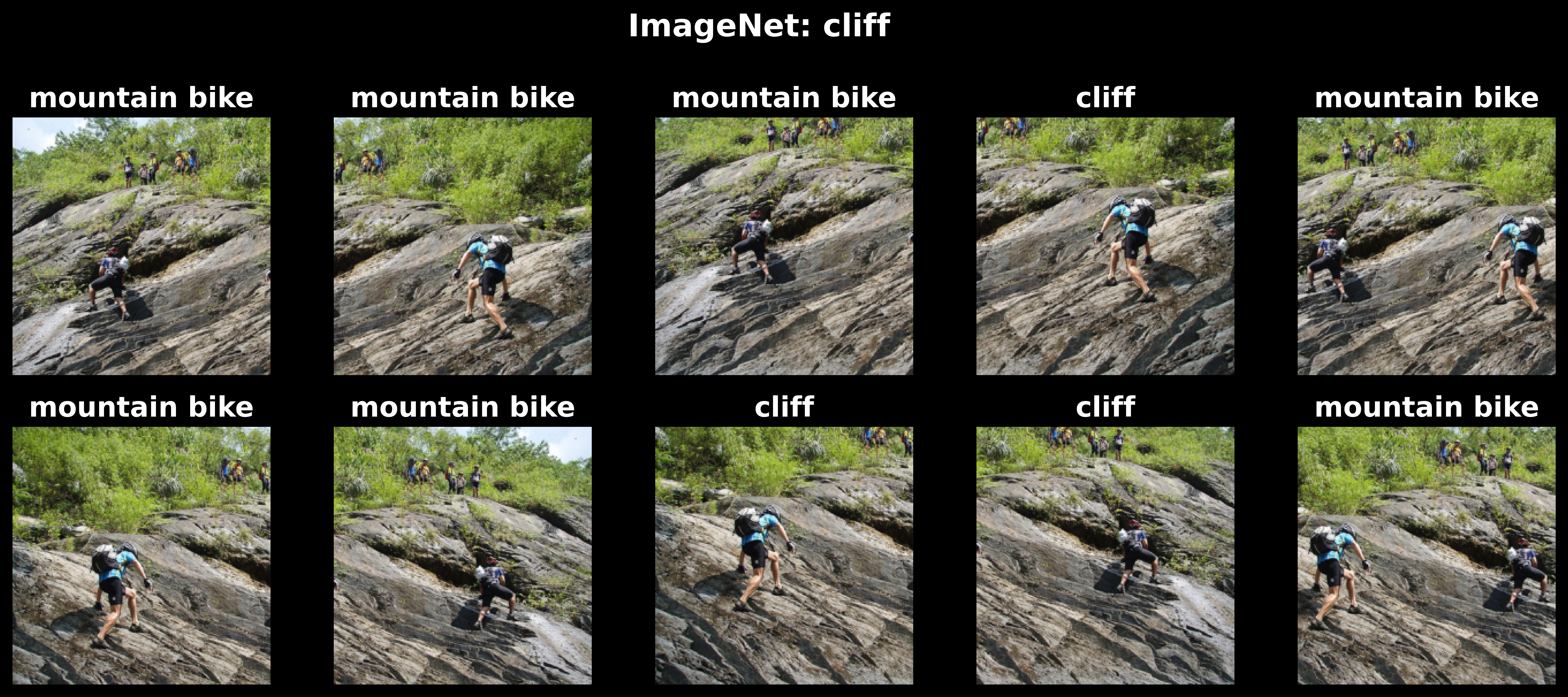}
        \label{ll3}
    \end{subfigure}
\caption{
Jittering of supervised ResNet50 prediction depending on a random sample of augmentation.
}
\label{fig:qwe1222}
\end{figure}

However, a network is only partially resistant to these transformations after training. 
Predictions can jitter depending on small changes in data (here we mean data augmentation, not any kind of adversarial attack), that especially pronounced on visually similar classes (Fig. \ref{fig:qwe1222}). 
This effect might be caused by unstable decision boundaries.

The performance improvement obtained by averaging embeddings is, likely, caused by a step back from unstable borders in a local region of more stable predictions.

\textbf{1-d loss interpolations~} 

\begin{figure}[t!]
\centering
    \begin{subfigure}{1\linewidth}
    \centering
        \includegraphics[width=0.915\linewidth]{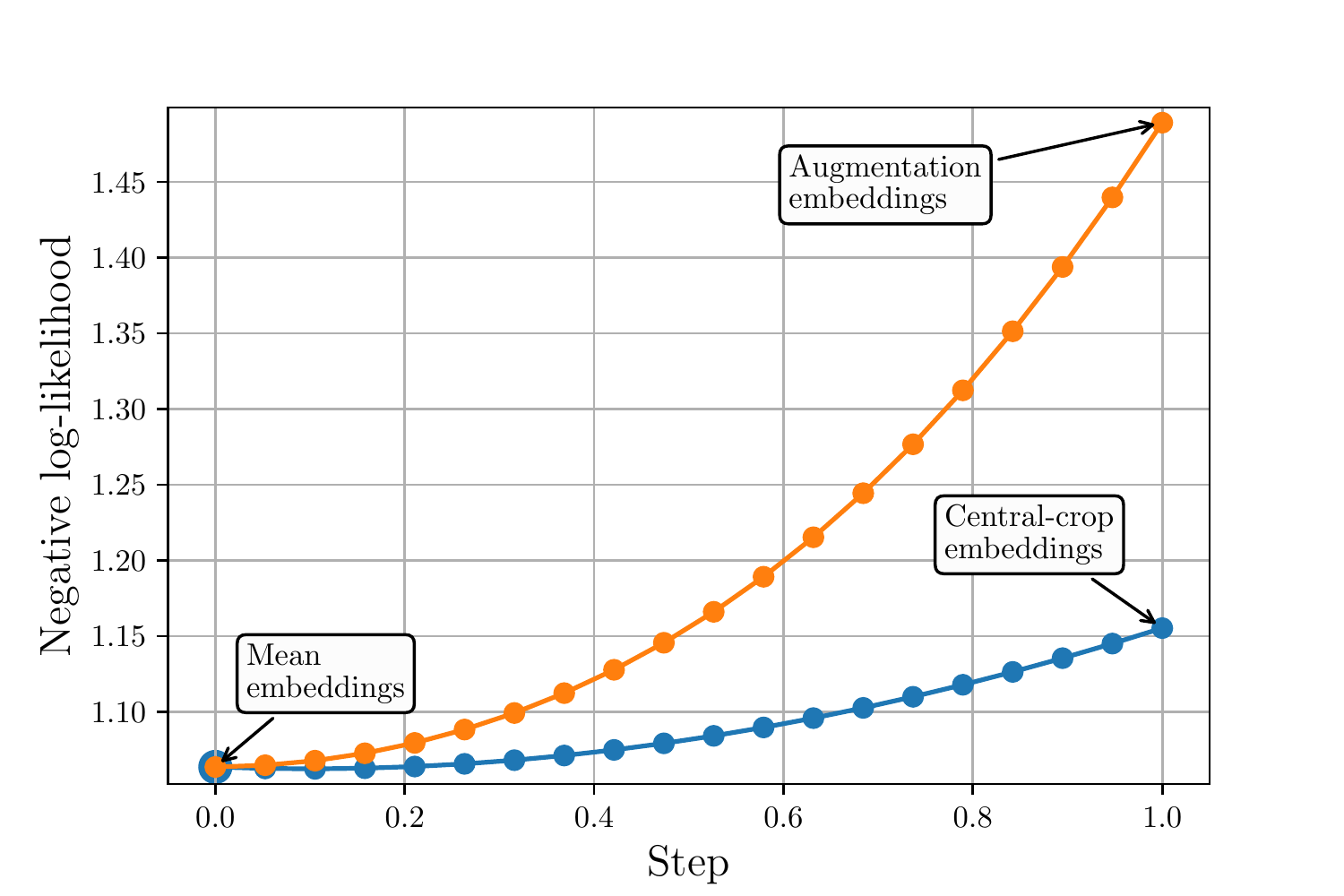}
        \label{ll3}
    \end{subfigure}
    
    \begin{subfigure}{1\linewidth}
    \centering
        \includegraphics[width=0.915\linewidth]{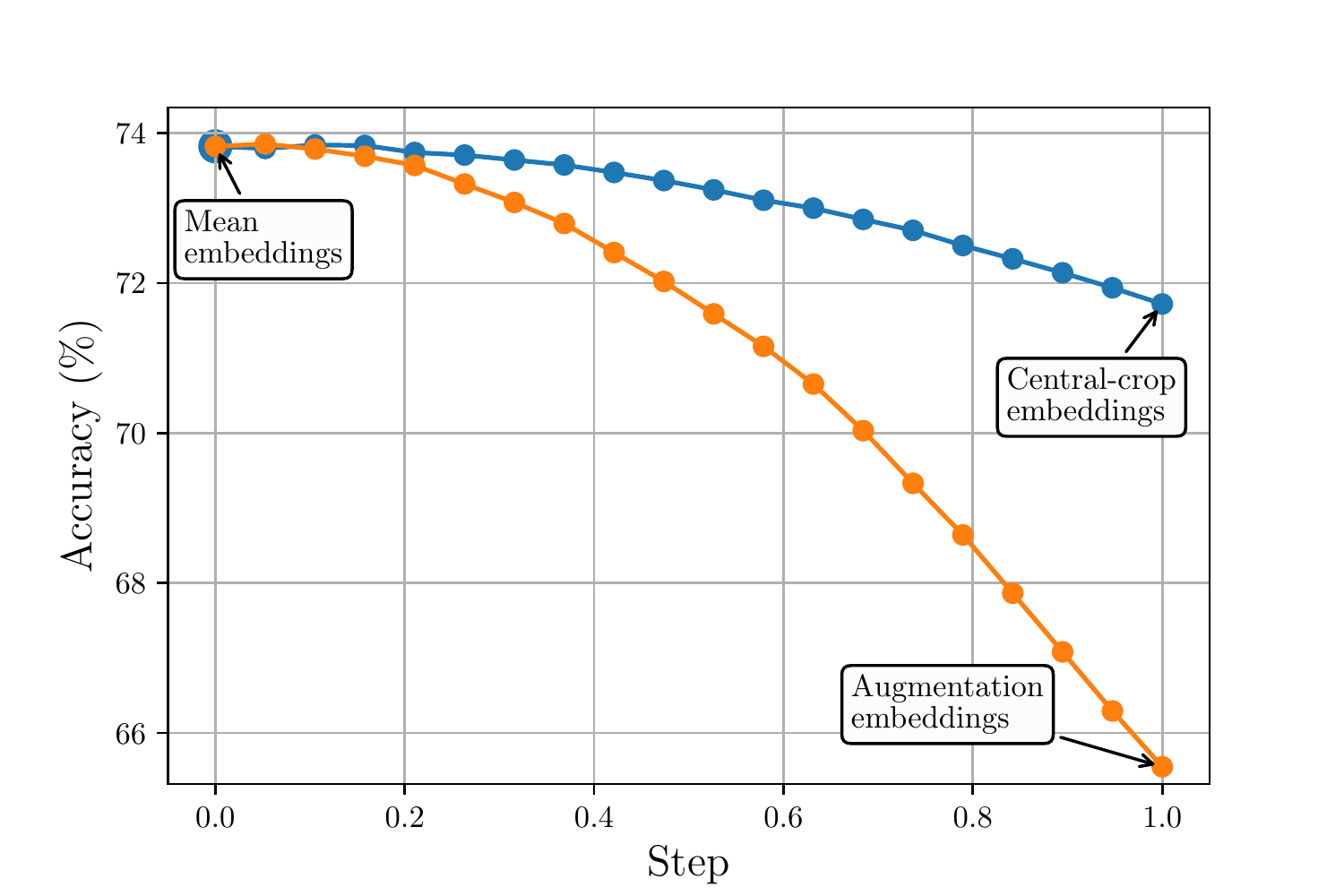}
        \label{ll3}
    \end{subfigure}
\caption{
Negative log-likelihood (top) and accuracy (bottom) for linearly interpolated embeddings of the form $(1-\alpha) \cdot x + \alpha \cdot y$, where $x$ is the mean embedding, and $y$ is the central-crop embedding for blue and an embeddings of an individual augmentation for orange.
Metrics for each step size $\alpha$ are averaged over validation images for both curves and additionally over different augmentations for the orange one.
The embeddings are taken from the ResNet50 (1$\times$, without SK) trained with SimCLRv2.
}
\label{fig:qwe111}
\end{figure}

Fig. \ref{fig:qwe111} shows that mean embeddings perform better than central crop embeddings and embeddings of individual augmentations.
It can be seen that loss decreases on average when approaching mean embeedings stating that this ME-direction in space of embedding is favorable.

\textbf{Relation to TTA \& pre-softmax averaging} 
Conventional test-time augmentation (eq. \ref{tta}) and mean embeddings with TTA (eq. \ref{metta}) are extremely similar:
\begin{gather}
    p(y\,|\,x)\!=\!\tcbhighmath{\E_{\hat{x} \sim p_{aug}(\cdot \,|\,x)}} \textsl{sm}(\theta^T a^L(\hat{x}; \mathrm{w})) \label{tta}\\
    p(y\,|\,x)\!=\!\textsl{sm}(\theta^T \tcbhighmath{\E_{\hat{x} \sim p_{aug}(\cdot \,|\,x)}} a^L(\hat{x}; \mathrm{w}))\label{metta}
\end{gather}
% Also, it is quite common that instead of averaging probabilities (eq. \ref{tta}), researches average activations of the last linear layer. 
It is quite common that researches average activations of the last linear layer instead of probabilities (eq. \ref{tta}). 
This, in the case of averaging feature from the last pre-softmax layer, is equivalent to MeTTA in terms of predictions:
\begin{multline}
    p(y\,|\,x)\!=\!\textsl{sm}(\theta^T \tcbhighmath{\E_{\hat{x} \sim p_{aug}(\cdot \,|\,x)}} a^L(\hat{x}; \mathrm{w})) = \\
    = \!\textsl{sm}(\tcbhighmath{\E_{\hat{x} \sim p_{aug}(\cdot \,|\,x)}} \theta^T a^L(\hat{x}; \mathrm{w})).
\end{multline}
An important difference is that MeTTA allows us to improve not only the final predictions but also representations, that can widen the scope of applications for ensembling techniques. 

The evaluation on a classification task serves as a sanity check for the MeTTA, as the regular TTA  performs~(roughly) the same. 
But, as we will highlight in the next section there are successful use cases of MeTTA-like method in image retrieval. 
% It is to be evaluated further in applications that rely on real-valued embeddings rather than class predictions.

\section{Related Work}

% The main goal of this work is to bring a new and hopefully wide area of applications to the light of the community of ensembling and uncertainty estimation. 
% The main goal of this work is to open a new and, hopefully, \textit{long-ranging} direction of the representation ensembling.
The main goal of this work is to draw the attention of ensembling and uncertainty estimation communities to the research on ensembling for enhancing the quality of representations.
% to improve the performance of \textit{their} methods. 
% new and hopefully wide area of applications to the light of the community of ensembling and uncertainty estimation. 

Ensembles of representations have been used for image retrieval for a while. Specifically, aggregated multi-scale representations, is used to average representations of an image at multiple scales \cite{gordo2017end, radenovic2018fine}\footnote{Code that uses multi-scale representations for image retrival {\small \url{https://github.com/filipradenovic/cnnimageretrieval-pytorch/blob/master/cirtorch/examples/test.py\#L53}}}.
Aggregated multi-scale representations is the successful use-cases of using a MeTTA-like method.
It also shows that we might need different augmentation policies depending on a specific application.

Invariant DNNs heavily exploit the idea of using data augmentations to develop invariant representations.
Specifically, TI-Pooling  \cite{laptev2016ti} computes the network predictions for multiple data augmentations in order to pool transformation-invariant features. 
The difference is that it does so during both training and inference, which makes its training step more expensive compared to MeTTA.

Independently, the idea of averaging representations has been proposed in Section 4 of \cite{foster2020improving}. 
The proposed methods are identical,
and share some motivation points e.g., a \emph{transformation invariant} propriety.
Theoretical explanations differ, and are complementary to each other e.g., we used lower-bound to motivate not using MeTTA during training, while \citet{foster2020improving} provides a nice intuitive motivation.
This work also contains experiments with large-scale self-supervised models.

\section{Conclusion and Open Directions}

We introduce MeTTA, a technique that uses a test-time augmentation ensemble in order to improve the representations quality. 
MeTTA improves the performance of both supervised and self-supervised features evaluated with linear evaluation benchmark.  
We also find that similar techniques have been used in image retrieval~\cite{gordo2017end, radenovic2018fine}, as well as similar ideas were used to learn feature invariant networks \cite{laptev2016ti}. 

MeTTA-like methods can be potentially applied to many problems, but one should be aware of the following pitfalls:
\begin{itemize}[noitemsep,nolistsep]
    \vspace{-\topsep}
    \item[$i)$] MeTTA for spatial features, that occur in problems like detection or segmentation, needs special averaging that accounts for offsets of crops and other deformations;
    \vspace{6pt}
    \item[$ii)$] The conventional data augmentation will not, most likely, fit for any problem. For example, we found that the conventional resize-randomcrop-flip augmentation hurts performance of image retrieval systems, whereas a widely used handcrafted multi-scale "augmentation" improves it. In general, the issue can be resolved with a policy search for (test-time) data augmentation~\cite{lim2019fast, molchanov2020greedy,kim2020learning, shanmugam2020and}. 
\end{itemize}
% \vspace{-0.45cm}
% What are the perspectives of using several models for inference better embeddings in deep ensemble like way?
There are the following perspectives for the usage of deep ensembles to inference better embeddings:
\begin{itemize}[noitemsep,nolistsep]
    \vspace{-\topsep}
    \item[$i)$] One way is to {\sl synchronize} embedding spaces between different models. As all predictions in deep ensembles are synchronized by the same ground truth, so we can average predictions. There are many known tricks that can be used for synchronization, e.g., contrastive losses~\cite{chen2020big, he2020momentum}. This is still an open direction. 
    \vspace{6pt}
    \item[$ii)$] The another way is to construct or learn an aggregation function for embeddings form non-synchronized networks. This, however, will most likely require the use of additional data, piece-wise training, as well as a the smart design of the function.
\end{itemize}

Ti-Pooling can be recognized to do both $i)$ \& $ii)$, but it shares the same weights across all models and train networks jointly which may hurt the predictive performance \cite{havasi2020training}.

During a review we received the following question: {\sl "Instead of only taking the mean of the embeddings, do you think it could be useful to also utilize other statistics (e.g., the variance)? Does the variance of the embedding relate to the uncertainty of the prediction?"}. We think it is a nice idea! It might be the way to introduce uncertainty to many problems.

We believe that spreading the success of ensembles to inference higher-quality representations is important, and will allow many new applications of ensembling.
MeTTA provides a small step in this exciting direction.

\bibliography{example_paper}
\bibliographystyle{icml2021}

\end{document}